\documentclass[12pt]{article}

\usepackage{amsmath}
\usepackage{amssymb}
\usepackage{mathtools}
\usepackage{amsthm}
\usepackage{graphicx}
\usepackage{a4}
\usepackage{parskip}
\usepackage{float}
\usepackage{hyperref}
\usepackage{xcolor}
\usepackage{color}
\usepackage{verbatim}
\usepackage{fancyhdr}
\usepackage[margin=1in]{geometry}

\title{Two Way Adversarial Unsupervised Word Translation}
\date{}

\begin{document}
\begin{center}	
	{\LARGE \bf{Two Way Adversarial Unsupervised Word Translation}}\\

	{\Large Blaine Cole}
\end{center}

\section{Introduction}

Word translation is a problem in machine translation that seeks to build models that recover word level correspondence between languages. Recent approaches to this problem have shown that word translation models can learned with very small seeding dictionaries, and even without any starting supervision. In this paper we propose a method to jointly find translations between a pair of languages. Not only does our method learn translations in both directions but it improves accuracy of those translations over past methods.

\section{Background}

\subsection{Formulation of the Problem}

The supervised word translation problem can be viewed as the orthogonal Procrustes problem:
\begin{align}
\min_{W\in O_n} \lvert\lvert WX - Y \rvert\rvert_F
\end{align}

In this setting the $X$, and $Y$ are matrices whose columns are word embedding vectors. This problem has an analytic solution shown in (2) if the columns of $X$ and $Y$ are aligned such that the word embeddings of words in language $X$ are at the same indices as their translations in language $Y$. 
\begin{align}
YX^T = U\Sigma V^T \Rightarrow W_{opt} = UV^T
\end{align}

Supervised and semi-supervised approaches to word translation have used the strategy of solving a Procrustes problem on a translated subset of words, then applying this solution to the rest of the language. These subsets have ranged from as large as 5000 translated pairs in Mikolov et al. 2013, to small bilingual dictionaries of only 25 pairs and potentially only paired translations of numbers in Artexe et al. 2017.

While the supervised approaches all assume knowledge of ground truth translations of some of the words, in the case that no translations are known, a more general view of the problem is:
\begin{align}
\min_{W\in O_n, \Pi} \lvert\lvert WX\Pi - Y \rvert\rvert_F
\end{align}

Where $\Pi$ is a permutation matrix that encodes the word level translation of these languages. As orthogonal Procrustes is efficiently solvable, unsupervised approaches to word translation first focus on finding an approximation to $\Pi$, then use that approximation to solve the Procrustes problem for $W$.

Conneau et al. 2018 approaches this problem by first finding an intermediary value of $W$ using the adversarial training procedure described in Goodfellow et al. 2014. Each batch samples of words from each language are drawn, source language words a put through a mapping network, while target language words are kept constant. A discriminator network then predicts the language that each word vector is associated with. Update steps alternate between training the discriminator to successfully predict the true language, and training the mapping to confuse the discriminator.

\subsection{Inference}

Given a mapping $W$ between the word embeddings of a source language and a target language, recovering a translation of a word is a non-trivial problem. The simplest method is to find nearest neighbors of the mapped source language words, in the target language, by looking at euclidean inner product of the translated word and each candidate word. This method has problems in higher dimensional spaces, due to the hub spoke phenomenon described in Radovanovi{\'c} et al 2010. In high dimensions mutual nearest neighbors are less likely to exist; instead, some vectors are the nearest neighbor of many different vectors.

Conneau et al. 2018 propose a different measure of similarity: Cross-Domain Similarity Local Scaling (CSLS). CSLS scores highly pairs of words which are similar to each other, but penalizes high average similarity with words in a neighborhood around the candidate word. The neighborhood has a fixed size and its members are the nearest neighbors of a  word in its given embedding space. For a word in the source language $x_i$, a potential translation in the target language $y_j$, the neighborhood $N(Wx_i)$ of size $T$ of mapped words around $Wx_i$, and $N(y_j)$ similarly being the neighborhood of target language words around $y_j$, CSLS is:
\begin{align}
	CSLS(Wx_i, y_j) = 2\langle Wx_i, y_j\rangle - \frac{1}{T} \sum_{y_t \in N(Wx_i)}  \langle y_t, Wx_i \rangle - \frac{1}{T} \sum_{Wx_t \in N(y_j)} \langle Wx_t, y_j \rangle
\end{align}

Other proposed similarity methods include Inverted Softmax, from Smith et al. 2017, and Hubless Nearest Neighbor Search, from Huang et al. 2019. Both of these methods require hyperparameter tuning to be optimal, which makes them better suited for use in a supervised setting where a seeding dictionary is available to validate hyperparameters. 

\subsection{Procrustes Refinement}

With the mapping $W$ learned from adversarial training an estimate for $\Pi$ is formed by finding pairs of mapped source language and target language words that are mutual nearest neighbors under CSLS similarity. For performance reasons the set of pairs searched over are limited to some subset of the most common words from the source language. Using this estimate of $\Pi$ we can solve (3) for $W$ and then perform translation with the refined mapping.

\section{Implementation}

\subsection{The Two Way Problem}

To improve on the standard adversarial method for word translation, we propose a new method for finding the intermediary $W$. Similar to Conneau et al. 2018 we find an initial $W$ through adversarial training and use it to perform the Procrustes refinement procedure described in section 2.3. Additionally we also simultaneously learn an additional mapping $Z$, that inverts $W$. More formally, if we consider a source language whose embeddings are denoted by $S$, and a target language whose embeddings are denoted by $T$, we define $W$ and $Z$ by equations (5) and (6).
\begin{align}
	W: S \rightarrow \hat{T}\\
	Z: \hat{T} \rightarrow \hat{S}
\end{align}

Note: we use $\hat{T}$, and $\hat{S}$ to indicate that these are approximations to the true word embedding spaces. 

To train these both adversarially we use two discriminators: $D_1$ which is trained to distinguish between $\hat{T}$ and $T$, and $D_2$ which is trained to distinguish between $\hat{S}$ and $S$. The mappings are then trained to confuse the discriminators. While $Z$ is trained to confuse $D_2$, $W$ is trained alternately to confuse $D_1$ then $D_2$. The standard adversarial losses are used to train the discriminators and mappings. For example for $D_1$ and $W$:
\begin{align*}
	\mathcal{L}_{D_1} = -\mathbb{E}_{x_s\sim S}\log D_1(Wx_s) - \mathbb{E}_{x_t\sim T}\log(1 - D_1(x_t))\\
	\mathcal{L}_W = -\mathbb{E}_{x_s\sim S}\log(1- D_1(Wx_s)) - \mathbb{E}_{x_t\sim T}\log D_1(x_t)
\end{align*}

The losses relating to $D_2$ follow analogously aside from the mapping now essentially being a two layer network without a non-linear activation layer. Additionally all the samples are drawn from the source language in steps that relate to $D_2$.

The architecture of the discriminators follows the architecture used in Conneau et al. 2018, the discriminators are both two hidden layer MLPs, with hidden layer size 2048, and leaky ReLU activation functions. We use stochastic gradient descent to train each of the networks with an initial learning rate of 0.1.

We hypothesize that making updates to $W$ to confuse $D_2$ will provide regularization to the training algorithm. By making updates to $W$ to allow the inverse mapping $Z$ to have a good solution, we will avoid finding trivial solutions for $W$ which can confuse $D_1$ but are sub-optimal for use in translation tasks. Additionally if we assume $W$ successfully maps to the target language space, then a good inverse mapping should act also be usable to recover translations from the target language to the source language.

\subsection{Orthogonality}

As shown in Smith et al. 2017, an orthogonality constraint on the mapping $W$ improves ultimate translation performance. As with Conneau et al. 2018 we adopt the procedure from Cisse et al. 2017, of partially solving an the optimization problem with the objective:
\begin{align}
	f(W) = \frac{\beta}{2}\lvert\lvert W^TW - I \rvert\rvert_F^2
\end{align}

The gradient of the objective function in (7) is:
\begin{align}
	\nabla_W f(W) = \beta(WW^T - I)W
\end{align}

Cisse et al. 2017 claims that finding $W$ to minimize (7) is untenable and instead only a single update step is needed.
\begin{align}
	W_{(k+1)} = W_{(k)} - \nabla_Wf(W_{(k)}) = (1+\beta)W_{(k)} - \beta W_{(k)}W_{(k)}^TW_{(k)}
\end{align}

We also implement a new initialization procedure for the mappings $W$ and $Z$. Rather than initializing to random values, we initialize to random orthogonal matrices by performing the QR decomposition on a random matrix of independent Gaussian samples. In general this seems to provide a speed up in convergence which is valuable as the two way model has double the number of parameters as the one way adversarial model. 

\subsection{Performance Criterion}

As a stopping criterion we look at the mean similarity between the 10000 most common source language words and their translations as proposed by $W$ after each epoch. CSLS and Euclidean inner product give similar results in terms of how model performance varies by epoch. As with Conneau et al. 2018, we experiment with using decreases in mean similarity to decay the learning rate of the mapping optimizers, but find that the best performing model states occur before such decreases are observed, typically after 2 or 3 epochs. Instead of using conditional learning decay, we decay the learning rate by a factor of 0.95 after each epoch.

\subsection{Size of Procrustes Refinement}

When performing the Procrustes refinement procedure, the number of source language queries used seems to be a factor. In table 1 we present accuracies for varying number of most common words from the source language used in the refinement process. There seems to be a threshold level in size, below which the Procrustes refinement is too unstable to produce good results, but after which increases yield little improvement. All following results use a size of 10000 during the refinement procedure. 

\begin{table}[h]
	\begin{center}
		\begin{tabular}{l | c | c | c | c}
			Size of search space & 1000 & 10000 & 20000 & 50000\\\hline\hline
			fr-en accuracy & 75.27 & 84.20 & 84.53 & 84.27
		\end{tabular}
		\caption{There seems to be some threshold for the size of the Procrustes refinement}
	\end{center}
\end{table}

\section{Results}

In table 2 we present results from our two way adversarial word translation method. We show translation results from using both the forward mapping $W$ and the inverse mapping $Z$ both as direct translations and after using them to find a starting point for the Orthogonal Procrustes problem. As a baseline we use the one direction results from Conneau et al. 2018. All experiments were run using fastText word embeddings trained on the Wikipedia corpus. Translation accuracies were computed using the fastText bilingual test dictionaries.

\begin{table}[h]
	\begin{center}
		\begin{tabular}{l | c c | c c | c c}
			Method and Metric &  en-fr & fr-en  & en-de & de-en & en-ru & ru-en\\\hline\hline
			From Conneau et al. 2018&&&&&&\\\hline 
			Adversarial + Procrustes - CSLS & 82.3 & 82.1 & 74.0 & 72.2 & 44.0 & 59.1\\
			Supervised Procrustes - CSLS & 81.1 & 82.4 & 73.5 & 72.4 & 51.7 & \bf{63.7}\\\hline
			Results from using forward mapping&&&&&&\\\hline
			Adversarial Mapping - Euclidean IP & 76.93 & 69.67 &  75.47 & 60.00 & 48.4 & 43.27\\
			Adversarial Mapping - CSLS & 83.53 & 76.20 & 81.87 & 65.27 & 54.27 & 47.53\\
			Adversarial + Procrustes - CSLS & \bf{86.47} & 84.20 & \bf{85.07} & 73.53 & \bf{63.0} & 54.93\\\hline
			Results from using inverse mapping&&&&&&\\\hline
			Adversarial Mapping - Euclidean IP & 70.93 & 72.23 & 70.53 & 63.6 & 47.73 & 38.07\\
			Adversarial Mapping - CSLS & 78.60 & 80.60 & 76.07 & 68.53 & 53.73 & 43.47\\
			Adversarial + Procrustes - CSLS & 85.87 & \bf{85.47} & 84.47 & \bf{75.13} & 61.47 & 54.0
		\end{tabular}
		\caption{Two way methods tend to out perform past work}
	\end{center}
\end{table}

In general the added regularization from jointly training the inverse map leads to improved translation performance. We replicate Conneau et al. 2018's results that show CSLS performs better than Euclidean inner product as a measure similarity for recovering translations. Somewhat surprisingly is the fact that inverse mapping works so well for translation. While the primary intent of it was to be a regularizer, it makes sense that a successful mapping from a good approximation the target language space to the source language space, would work for translation. 

\subsection{Analysis of Errors}

The ability for the inverse mapping to work as a translation implies that the forward mapping is successful at mapping source word embeddings into the target word embedding space. However forward mappings at best have an error rate of 14\%. This raises the question of where the errors are coming from if the forward mapping is successful. Table 3 shows the first 8 errors from the French to English translation experiment.

\begin{table}[h]
	\begin{center}
		\begin{tabular}{c | c | c | c}
			Source word & Predicted Translation & Ground Truth Translation & Ground Truth Position\\\hline\hline
			sac & sack & bag & 2\\
			concern{\'e}s & relevant & concerned & 9\\
			hard & punk & hardcore & 2\\
			essor & burgeoning & booming & 5\\
			chasseur & hunters & fighter & 6\\
			null{\'e} & zero & null & 6\\
			mat{\'e}riau & tensile & material & 3\\
			l{\'e}gale & statutory & legal & 2
		\end{tabular}
		\caption{Translation errors are not egregiously off}
	\end{center}
\end{table}

While anecdotal these errors give some intuition as to the kind of mistakes the model makes. Some errors, such as sack vs bag, and zero vs null, would work as translations in many cases despite being incorrect. What these errors lose in translation is the nuance of the words. Other errors such as burgeoning vs booming, and statutory vs legal, still share semantic meaning to a degree which suggests that alignment of the embeddings has been successful, to the extent that semantic meaning has been encoded in the word embeddings.

\subsection{Visualization of Language Alignment}

To get a sense of the level of alignment we can look UMAP, McInnes et al. 2018, projections of the word vectors before and after the Procrustes mapping process.

\begin{figure}[h]
	\begin{center}
		\includegraphics[width=2.7in]{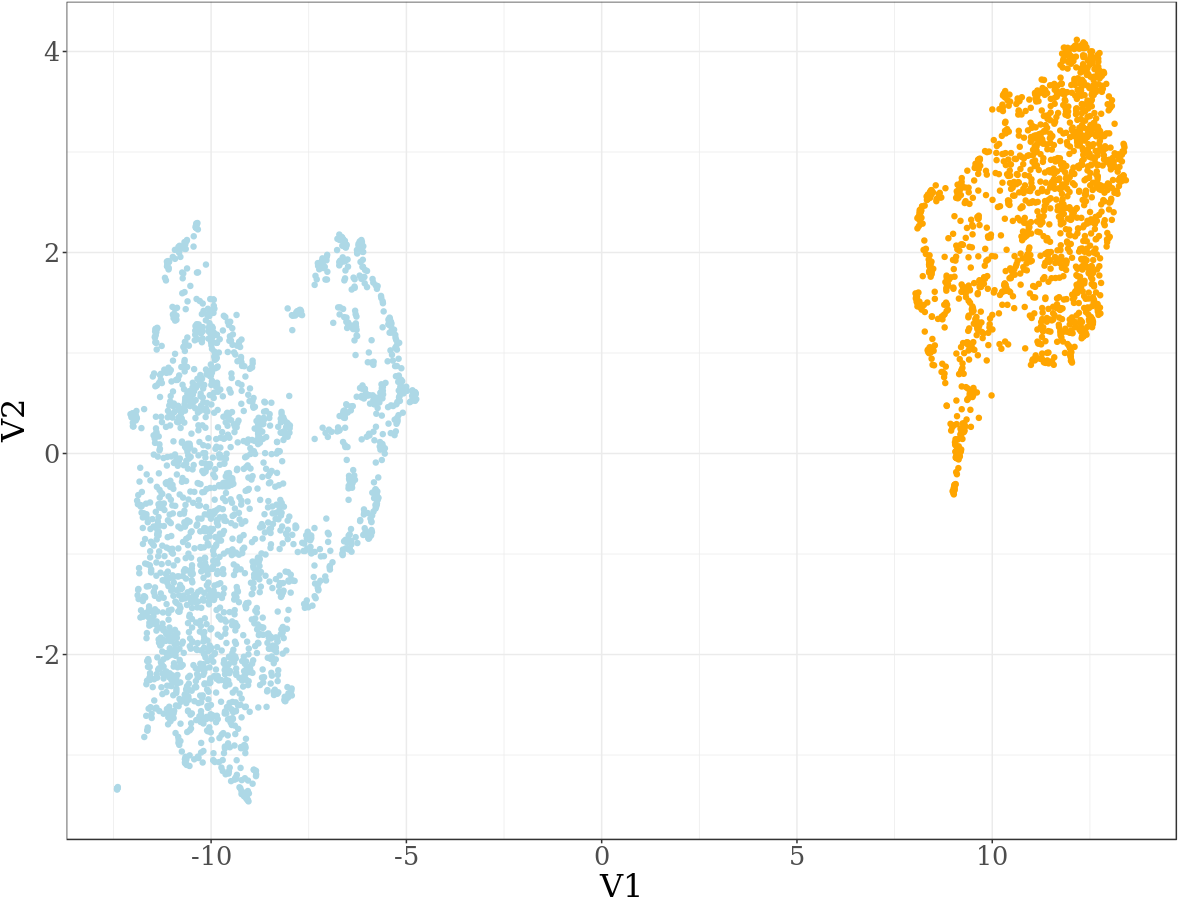}
		\includegraphics[width=3.2in]{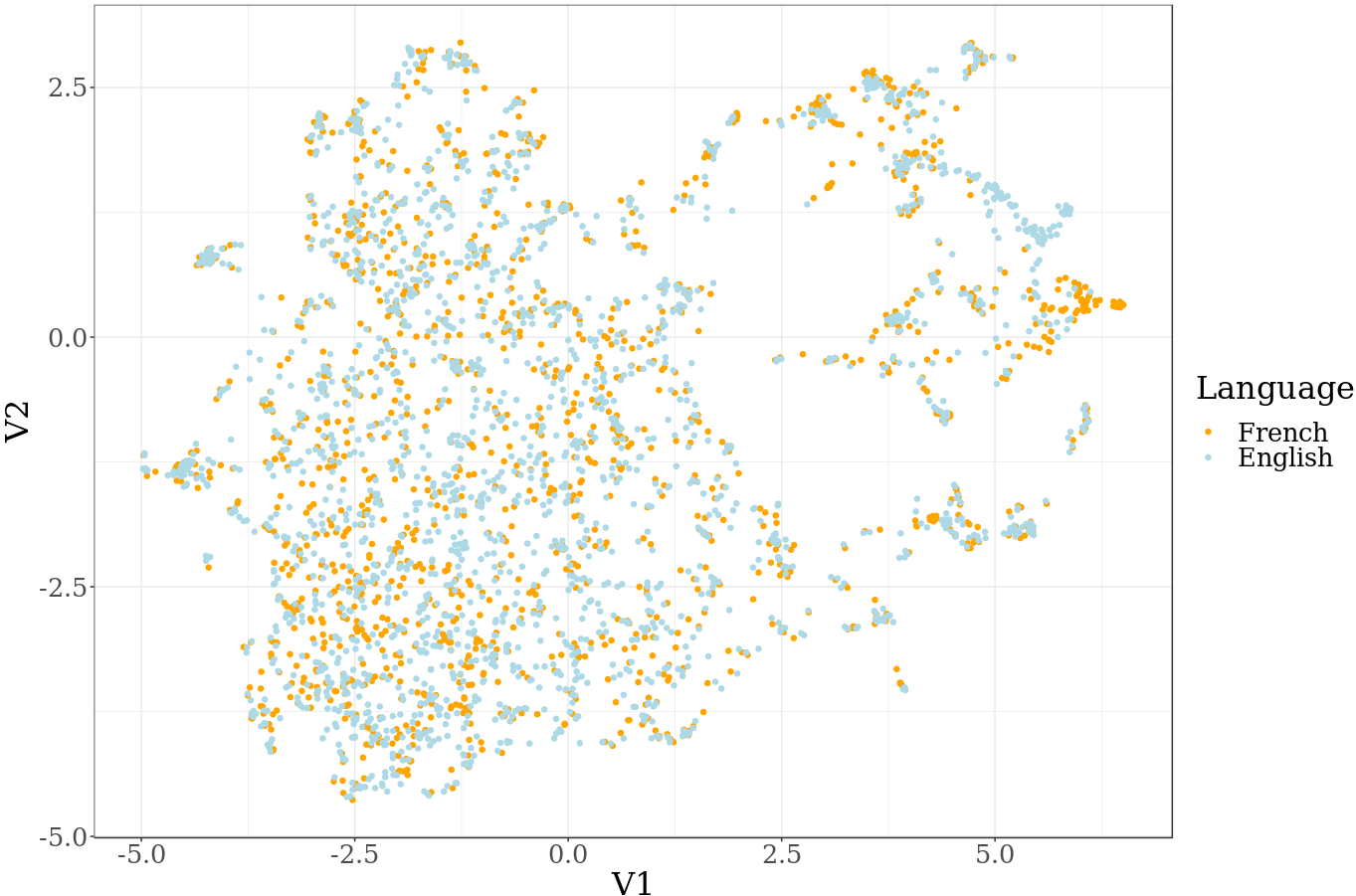}
		\caption{Left: 1500 query words from French and their ground truth English translations; Right: English word vectors + mapped French word vectors}
	\end{center}	
\end{figure}

Figure 1 gives a good illustration of the efficacy of the learned mapping. Before the mapping, the UMAP projection completely separates the French and English word vectors; however, after, the French and English word vectors are clearly aligned. 

\section{Conclusion}

In this paper we have proposed a method of jointly performing unsupervised word translation between two languages, that for most tasks out performs past supervised and unsupervised methods. The joint translation not only adds regularization and improves accuracy, but also allows us to find translations in both directions simultaneously. Additionally the ability to learn both translations provides gives substantial evidence that the embedding spaces are being effectively aligned.


\begin{thebibliography}{9}
	\bibitem{mikolov}
	Tomas Mikolov, Quoc V Le, Ilya Sutskever. Exploiting Similarities among Languages for Machine Translation. 2013
	\bibitem{artexe}
	Mikel Artexe, Gorka Labaka, Eneko Agirre. Learning Bilingual Word Embeddings with (Almost) no Bilingual Data. Proceedings of ACL 2017
	\bibitem{conneau}
	Alexis Conneau, Guillaume Lample, Marc'Aurelio Ranzato, Ludovic Denoyer, Herv{\'e} J{\'e}gou. Word Translation without Parallel Data. Proceedings of ICLR, 2018.
	\bibitem{good}
	Ian Goodfellow, Jean Pouget-Abadie, Mehdi Mirza, Bing Xu, David Warde-Farley, Sherjil Ozair, Aaron Courville, Yoshua Bengio. Generative Adversarial Networks. Advances in Neural Information Processing Systems, 2014.
	\bibitem{rad}
	Milo{\v s} Radovanovi\'{c}, Alexandros Nanopoulos, and Mirjana Ivanovi\'{c}. Hubs in Space: Popular Nearest Neighbors in High-Dimensional Data. Journal of Machine Learning Research 2010.
	\bibitem{smith}
	Samuel L. Smith, David H.P. Turban, Steve Hamblin, and Nils Y. Hammerla. Offline Bilingual Word Vectors, Orthogonal Transformations and the Inverted Softmax. ICLR 2017.
	\bibitem{huang}
	Jiaji Huang, Qiang Qiu, Kenneth Church. Hubless Nearest Neighbor Search for Bilingual Lexicon Induction. ACL 2019.
	\bibitem{cisse}
	Moustapha Cisse, Piotr Bojanowski, Edouard Grave, Yann Dauphin, Nicolas Usunier. Parseval Networks: Improving Robustness to Adversarial Examples. ICML 2017.
	\bibitem{umap}
	Lelang McInnes, John Healy, James Melville. UMAP: Uniform Manifold Approximation and Projection for Dimension Reduction. 2018.
	
\end{thebibliography}
\end{document}